# Paddy Disease Detection and Classification Using Computer Vision Techniques: A Mobile Application to Detect Paddy Disease

Bimarsha Khanal
*Department of Electronics and Computer Engineering,
Paschimanchal Campus,
Tribhuvan University*
Pokhara, Nepal

Paras Poudel
*Department of Electronics and Computer Engineering,
Paschimanchal Campus,
Tribhuvan University*
Pokhara, Nepal

Anish Chapagai
*Department of Electronics and Computer Engineering,
Paschimanchal Campus,
Tribhuvan University*
Pokhara, Nepal

Bijan Regmi
*Department of Electronics and Computer Engineering,
Paschimanchal Campus,
Tribhuvan University*
Pokhara, Nepal

Sitaram Pokhrel
*Department of Electronics and Computer Engineering,
Paschimanchal Campus,
Tribhuvan University*
Pokhara, Nepal

Salik Ram Khanal
*School of Business and Technology, Curry College*,
Milton, MA, USA

## Abstract

Plant diseases significantly impact our food supply, causing problems for farmers, economies reliant on agriculture, and global food security. Accurate and timely plant disease diagnosis is crucial for effective treatment and minimizing yield losses. Despite advancements in agricultural technology, a precise and early diagnosis remains a challenge, especially in underdeveloped regions where agriculture is crucial and agricultural experts are scarce. However, adopting Deep Learning applications can assist in accurately identifying diseases without needing plant pathologists. In this study, the effectiveness of various computer vision models for detecting paddy diseases is evaluated and proposed the best deep learning-based disease detection system. Both classification and detection using the *Paddy Doctor* dataset, which contains over 20,000 annotated images of paddy leaves for disease diagnosis are tested and evaluated. For detection, we utilized the YOLOv8 model-based model were used for paddy disease detection and CNN models and the Vision Transformer were used for disease classification. The average mAP50 of 69% for detection tasks was achieved and the Vision Transformer classification accuracy was 99.38%. It was found that detection models are effective at identifying multiple diseases simultaneously with less computing power, whereas classification models, though computationally expensive, exhibit better performance for classifying single diseases. Additionally, a mobile application was developed to enable farmers to identify paddy diseases instantly. Experiments with the app showed encouraging results in utilizing the trained models for both disease classification and treatment guidance.

## Introduction

Paddy, also known as rice, is a major food staple for nearly half the world's population. It is particularly vital in Asian countries, where it is consumed and grown extensively. India and China are the leading producers, contributing to approximately 50% of global rice production [1]. The agricultural economies of these nations heavily depend on rice cultivation, making it crucial to ensure a good yield of paddy. In Asia, rice is a major food source, delivering up to 50% of the dietary calories for millions living in poverty, and is essential for maintaining food security [2]. Crop disease significantly reduces yields by affecting various aspects of food production such as overall production, physical availability, distribution, economic access, stability, quality, and nutritional value. Paddy also faces several diseases that adversely impact yields, leading to yield losses of up to 30% and posing considerable challenges in maintaining sufficient production [3]. Various factors contribute to these diseases, including viral, bacterial, and fungal infections, pest infestations, and physical damage. Blast, blight, brown spot, stem borer, tungro virus, downy mildew, and insect infestation are some major diseases prevailing in paddy crops. Nepal alone has around 20 identified paddy diseases, highlighting the diversity of challenges farmers face in developing countries [4]. Early diagnosis and treatment of these diseases are crucial to minimize their effects and prevent significant yield loss and economic losses for farmers.

Detecting and classifying diseases in plants, especially paddy being a primary food source, requires a lot of knowledge that may not be universally available. In most underdeveloped and developing countries, especially in Asia, there is a lack of sufficient knowledge about diseases among farmers, and there is a shortage of agricultural pathologists to provide expert guidance. Moreover, the similar appearance of many paddy diseases makes manual identification challenging. Farmers also require chemical pesticides to treat paddy diseases, and it is crucial for them to use the correct types of chemicals according to the specific disease and to apply them as early as possible to prevent yield loss [5]. Misidentification or delayed treatment can lead to the use of incorrect pesticides, which not only fail to address the disease but can also cause harm to the crops and the environment. Therefore, developing an easily accessible automated system using advanced deep learning techniques can greatly benefit farmers by providing quick and reliable disease diagnosis, ultimately improving crop management and yield.

The field of plant disease detection has witnessed substantial progress by applying digital image processing and machine learning techniques. Researchers have investigated diverse methodologies, encompassing Deep Convolutional Neural Networks (DCNN), Support Vector Machines (SVM), and region-based CNNs. Notwithstanding these advancements, significant challenges persist. These include the complexities associated with disease localization in natural environments, the computational demands of current models, and the susceptibility to overfitting in scenarios with limited datasets. Moreover, technical hurdles such as low image contrast, variability in lesion area scales, and the presence of noise in captured images continue to impede the efficacy of existing solutions. This persistent research gap underscores the imperative for developing more robust, computationally efficient, and accurate plant disease detection systems. Such advancements are crucial for overcoming current limitations and delivering practical solutions that can contribute meaningfully to global food security initiatives.

Several notable challenges have been identified after a thorough study of plant disease identification approaches. One of the primary issues is localizing diseased regions within images, particularly in complex natural environments where backgrounds can be cluttered and indistinct. Additionally, many models are computationally heavy, making them less suitable for real-time accurate detection. The challenge is further increased by the risk of overfitting, particularly when dealing with smaller datasets. The minute difference between lesion area and background, low contrast, significant scale variations in the lesion, and noise on the lesion are the other challenges found in a complex natural environment.

In this study, we propose to explore various deep-learning approaches for the identification of paddy diseases, focusing on both detection and classification tasks. We will compare the performance of several state-of-the-art models, including Convolutional Neural Networks (CNNs), Vision Transformers, and YOLOv8 for their effectiveness in accurately identifying and classifying paddy diseases. The study aims to not only evaluate these models but also develop a user-friendly mobile application that can automate the process of disease detection and provide timely treatment suggestions to farmers. This application is intended to bridge the knowledge gap in disease identification, enabling farmers to take appropriate and timely actions to manage paddy diseases effectively, ultimately contributing to improved crop yields and food security.

# Related Work

Before the advent of modern technology, plant disease diagnosis was primarily done through visual inspection by experts. Farmers would rely on their own experience and knowledge to identify diseases based on the appearance of leaves, stems, and other plant parts. In addition to visual inspection, various lab tests have traditionally been employed to confirm the presence of diseases in paddy plants. These manual approaches were time-consuming, subjective, and often required specialized expertise that was not readily available to all farmers, especially in developing regions [6].

With the advancements in computational technology, the adoption of machine learning and digital image processing, particularly since the late 1990s and early 2000s, has significantly transformed plant disease detection. An expert system-based prototype system for diagnosing paddy diseases like Blast, Brown Spot, and Narrow Brown Spot achieved 94.7% accuracy by extracting features such as lesion type and color using thresholding and morphological algorithms, highlighting its potential for further improvement [7]. Similarly, segmentation methods such as k-means, thresholding, Otsu's method, and Fuzzy C-means, an extension of k-means for handling overlapping clusters have been widely applied. For feature extraction, techniques like color extraction, edge detection, color coherence methods, and shape recognition have been used to analyze visual patterns in diseased areas. These extracted features are then classified using machine learning algorithms such as Support Vector Machines (SVM), k-nearest Neighbors (k-NN), and simple neural networks, providing accurate and computationally efficient solutions [8], [9], [10] .

In recent years, deep learning has gained prominence in plant disease identification, leveraging convolutional neural networks (CNNs) and other architectures to achieve superior performance over traditional methods. Nalini [11] proposed a deep neural network (DNN) model optimized with a crow search algorithm (CSA), achieving superior accuracy compared to machine learning models like SVM. The approach involved preprocessing plant images with k-means clustering and thresholding, followed by feature extraction and classification, demonstrating reduced computational workload and high efficiency. Ahmed [12] proposed an Efficient Deep Learning-based Fusion Model that combined traditional image processing techniques with deep learning to achieve accurate disease detection and classification, with a reported accuracy of 96.17%. The method employed median filtering and k-means segmentation for preprocessing, integrated features from the Gray Level Co-occurrence Matrix (GLCM) and Inception-based deep features, and utilized an optimized Fuzzy Support Vector Machine (FSVM) for classification, demonstrating superior performance over existing approaches.

In recent years, high-quality image datasets have become more accessible, encouraging the adoption of convolutional neural network (CNN) architectures for plant disease detection, which excel in image-based tasks. Sladojevic [13] developed a deep convolutional neural network (DCNN) model to classify fifteen different plant diseases, including healthy and background classes, achieving an average accuracy of 96.3%. Similarly, Alfarisy [14] used a CaffeNet model trained on a dataset of 4,511 images across thirteen classes of paddy pests and diseases, achieving an accuracy of 87%. Faez[15] experimented with five different deep learning models CNN, VGG16, VGG19, Xception, and ResNet50 found that Xception achieved the highest performance, with an accuracy of 98% for detecting paddy diseases. Similarly, Inception-V3 showed 96.23% accuracy in classifying paddy leaf diseases [16], while DenseNet outperformed other models with 99% accuracy in the study [17]. Additionally, a region-based CNN model demonstrated success in disease prediction for tomato plants, showing promise for scalable agricultural monitoring [18].

Recent research has focused on utilizing object detection techniques for the real-time identification of paddy leaf diseases [19]. An improved version of the YOLOv5 model was proposed by Gao [20] which achieved an mAP of 9.9% better accuracy than the original version of YOLOv5. The YOLOv7 model has demonstrated superior performance in rice leaf disease detection, achieving high precision and recall with metrics such as 90% precision and 81% F1 score, leveraging 1500 annotated data samples [21]. Additionally, the integration of pyramid YOLOv8 has enhanced disease detection accuracy, achieving an average mAP value of 0.84, making it a robust tool for timely disease prevention in rice farming [22]. Furthermore, the Faster R-CNN model, trained with both online and field datasets, exhibited exceptional accuracy rates for diagnosing rice blast, brown spot, and hispa, as well as healthy rice leaves, with accuracies surpassing 99%, highlighting its effectiveness in real-time disease detection [23]. These advancements in object detection, particularly through the YOLO and Faster R-CNN families, underscore the potential of deep learning-based systems to improve rice disease monitoring and contribute to sustainable agricultural practices.

Recently, Vision Transformer (ViT)--based models have become the popular option in plant leaf disease detection due to their superior results in visual recognition tasks outperforming CNN-based methods. For instance, [24] proposed the Plant Disease Localization and Classification model (PDLC-ViT), which integrates co-scale, co-attention, and cross-attention mechanisms within a Multi-Task Learning (MTL) framework to achieve remarkable accuracy in detecting and classifying plant diseases, reaching an accuracy of 99.97%. Another study introduced a hybrid model combining the feature extraction capabilities of CNNs with the strengths of ViT, achieving an accuracy of 98.86% on the Plant Village dataset, showcasing its effectiveness in real-world conditions [25]. Additionally, the application of ViT models to multispectral imaging for plant disease detection also demonstrated impressive results, with accuracies of 93.71% and 90.02% in real-world field conditions. These studies emphasize the growing significance of ViT models in advancing plant disease detection systems and their integration into agricultural practices.

Several mobile applications are available online claiming to detect different types of plant diseases, but only a few of them are effective. Plantix – Your Crop Doctor is among the most popular, capable of detecting common plant diseases (GmbH). Few studies have discussed the development of mobile applications designed to aid in paddy disease detection. Ng [26] introduced a mobile application using the Faster R-CNN object detection model with an Inception-v2 backbone, achieving 97.9% accuracy for grape disease detection on smartphones without server reliance. Another innovative solution is a LINE Bot system for diagnosing rice diseases in real-time, employing YOLOv3 and achieving an Average True Positive Point of 95.6% during testing and 78.86% in real-world deployment. This system provides farmers with automatic, fast, and actionable disease diagnoses directly through their devices [27] .

## Materials and Methods

### Data Collection and Preparation

The images used in this study were sourced from the Paddy Doctor (Murugan) dataset, a comprehensive collection specifically curated for paddy disease detection. The dataset comprises a total of 16,225 images categorized into 13 distinct classes, which include a normal paddy plants class and 12 disease classes (see Figure 1). The dataset covers a diverse range of paddy diseases which are Fungal (Brown Spot, Blast, and Downy mildew), Bacterial (Bacterial leaf streak, Bacterial leaf blight, and Bacterial panicle blight), Viral (Tungro), and Pests (Yellow stem borer, Black stem borer, Leaf roller, White stem borer, and Hispa). The images are of 1,080 x 1,440 pixels collected using a smartphone camera. The distribution of all types of images is shown in Figure 2. This dataset was utilized for both classification and detection tasks. For detection purposes, bounding boxes were manually drawn on 200 images of each class around the diseased areas with the guidance of domain experts. Label Studio was used for dataset labeling to ensure accurate annotation of disease-affected areas.

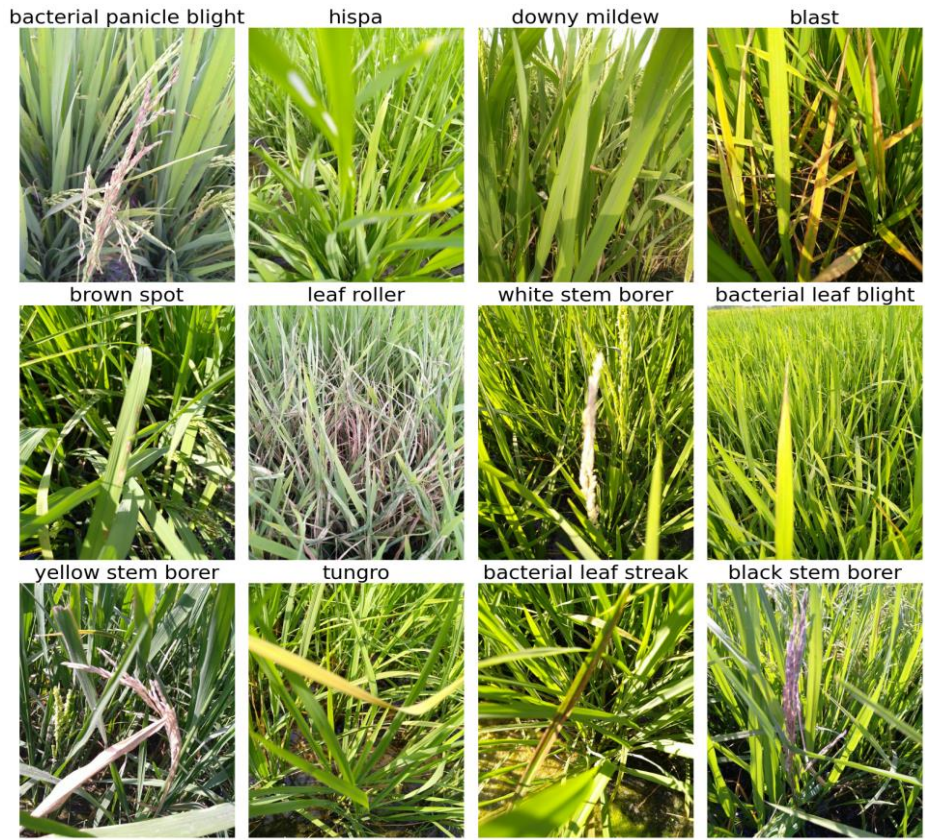

Figure 1: Sample images from Paddy Doctor Dataset

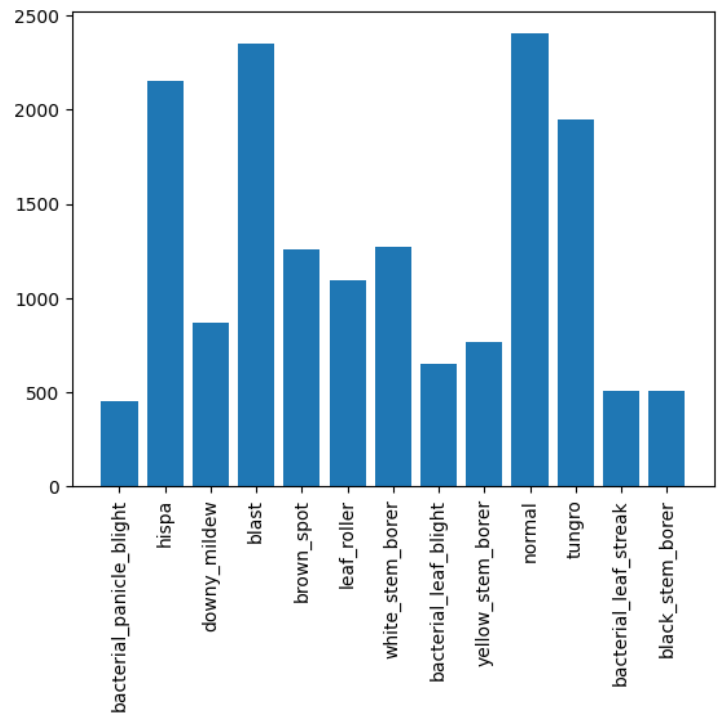

Figure 2: Image count per disease category in Paddy Doctor dataset for disease Identification Task

Image augmentation plays a crucial role in enhancing the robustness and generalization ability of deep learning models by artificially expanding the training dataset. In this study, image transformation techniques for augmentation, including rotation, flipping, random brightness adjustment, and shearing were used to enhance the diversity in training images. Augmentation steps were used both for classification and detection tasks. For detection, bounding boxes were also transformed with a minimum visibility of 30% for bounding box preservation after random geometric transformation. The data was split into training (80%) and testing (20%) sets before augmentation was applied to ensure sufficient data for model training and evaluation.

## Methodology

The overall prediction techniques are divided into classification and detection tasks. The detection task used deep learning-based object detection algorithms to detect various types of diseases whereas the classification task will be able to classify the type of disease. The basic block diagram of the paddy disease detection and classification methods is in Figure 3.

## Paddy Disease Classification

For paddy disease classification tasks, several experiments were carried out based on CNN and Vision Transformer architectures. The models were evaluated using a randomly selected validation set comprising 20% of the total images from the dataset.

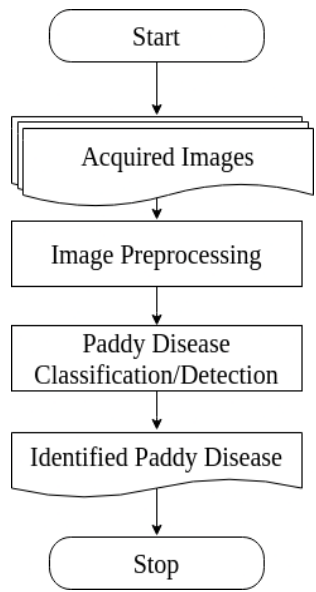

Figure 3: Overview of methodological steps

**CNN Architectures**

A CNN architecture is a linear or non-linear combination of convolutional layers that extract features, pooling layers that down-sample the data, activation functions that introduce non-linearity, and fully connected layers that combine features. It excels at recognizing patterns and extracting features from data with grid-like structures such as images, making it a powerful tool for various computer vision tasks like image classification, object detection, and image segmentation. ResNet, VGGNet, MobileNet, AlexNet, and EfficientNet are some of the popular CNN architectures.

CNN architectures are well-suited for paddy disease classification due to their ability to automatically learn and extract hierarchical features from images. They specialize in capturing spatial patterns and textures, which are crucial for identifying disease patches in paddy leaves. The convolutional layers effectively handle variations in scale, position, and orientation of the disease patches, leading to robust and accurate classification. Additionally, the use of pre-trained CNN models allows leveraging knowledge from large datasets, further enhancing their performance on varying paddy disease images.

**ResNet**

The general idea of "the deeper the better" holds for CNN architectures since the models' flexibility to adapt to any space increases because they have a bigger parameter space to explore. However, it is observed that performance degrades after a certain depth due to the vanishing gradient problem, where information essential for training earlier layers gets lost during backpropagation [28].

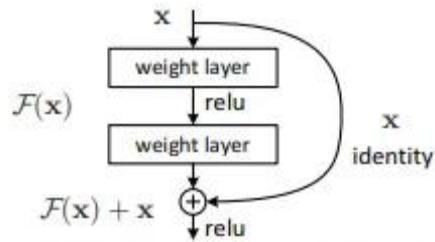

Figure 4 : Residual Block of Resnet [28].

Instead of learning a direct mapping from input to output, a residual block learns a residual function, i.e. the difference between the desired output and the input. This is achieved by using a "shortcut connection" that skips one or more layers within the block, adding the input to the output of the skipped layers. So, the skip function acts as an identity function. The output of the skip connection and weight layer are added together. The skip connections are passed through a CNN layer if the dimension of output from the weight layer and skip connection don't match [28]. The use of such skip connections supports the deep architecture of ResNet to capture detailed and hierarchical features of paddy disease symptoms, making it proficient at distinguishing between complex and similar-looking disease patches. The ability to identify subtle differences in texture, color, and shape is crucial for accurate classification of paddy diseases.

**Vision Transformer**

The Vision Transformer (ViT) is an emerging architecture that is outperforming CNN architectures for computer vision tasks. It utilizes the power of self-attention mechanisms inspired by transformers, originally designed for natural language processing tasks. Convolutional layers in traditional CNN are replaced by self-attention layers, enabling direct interactions between all input patches. This architecture excels at capturing global dependencies and contextual relationships within images, making it highly effective for tasks such as image classification, object detection, and image segmentation [29]. In our study, ViT has shown slightly superior performance compared to CNN architectures.

The first layer of Vision Transformer is the patch embedding layer which divides input images into fixed-size patches, treating each patch as a token like in language processing tasks. Each patch is linearly embedded into a lower-dimensional representation. This embedding step converts image patches into sequences of vectors, making them suitable for processing by the transformer layers.

The rest of the architecture is like the transformer architecture used for Natural Language Processing tasks.

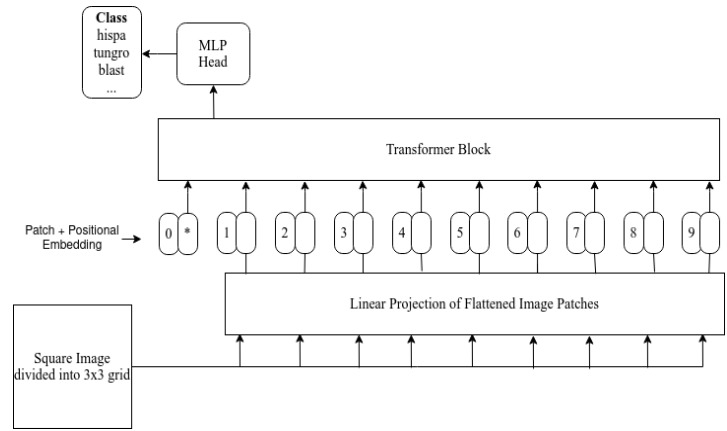

Figure 5: Vision Transformer architecture [29]

Unlike CNNs, Vision Transformers process images as sequences of patches, allowing them to focus on different parts of the image simultaneously and understand intricate patterns across the entire image. This global self-attention mechanism is particularly useful for identifying subtle and distributed disease symptoms in paddy leaves, though it comes at the cost of increased training and inference time.

In the paddy disease classification task, the categorical cross-entropy loss was used as a loss function for model optimization. The categorical cross-entropy loss function is defined in equation 1.

$$Loss_{classification} = -\frac{1}{N}\sum_{i=1}^{N}\sum_{c=1}^{C} y_{i,c} \log(\hat{y}_{i,c}) \dots\dots\dots\dots\dots[1]$$

Where,

C is the number of classes.

$y_{i,c}$ are the true labels for class c for instance i

$\hat{y}_{i,c}$ is the predicted probability for class c for instance i

The performance of all the classification models was evaluated using metrics such as accuracy, precision, recall, and F1-score, ensuring they meet the high standards required for practical agricultural applications.

## Paddy Disease Detection

For the paddy disease detection task, the study employs the YOLO (You Only Look Once) model, a CNN-based architecture for object detection. Unlike the classification task, YOLO is particularly advantageous when dealing with images containing multiple instances of different paddy diseases, as it processes the entire image in a single forward pass, enabling simultaneous detection of various objects with high precision.

**YOLO Architecture**

You Only Look Once is the CNN-based state-of-the-art real-time object detection algorithm designed for single-shot object detection and classification tasks. YOLO treats object detection as a single regression problem which makes it different from other window sliding approaches. For our experiment, we used the YOLOv8 model and compared the results with previous YOLO models. Some major changes in YOLOv8 are anchor-free classification heads and some tweaks in backbone and stem blocks. The use of anchor free head eliminates the need for predefined anchor boxes resulting in better accuracy whereas changes in backbone and stem block contribute to better inference speed [30].

**Components of YoloV8**

*Backbone*

The backbone is the CNN block responsible for feature extraction from the input image. CSPDarknet53 is used as the backbone in the YoloV8 model, which is built upon Darknet-53 architecture with an added Cross Stage Partial Network strategy to improve efficiency and accuracy. CSPNet splits the feature maps from CNN blocks and each split undergoes separate processing through a lightweight residual block. Finally, the processed features from both parts are concatenated. This crucial step allows information to flow across different stages of the network, facilitating better gradient propagation and feature representation.

*Neck*

The neck, which acts as the feature extractor, combines feature maps from different stages of the backbone to collect information across multiple scales. YOLOv8 introduces a novel C2F module different from the conventional Feature Pyramid Network (FPN). This module integrates high-level semantic features with low-level spatial details, resulting in enhanced detection accuracy, particularly for smaller objects.

*Head*

The head is the final stage of YOLov8 responsible for making predictions. Unlike its predecessors, YOLOv8 features multiple detection modules for predicting bounding boxes, objectless scores, and class probabilities for every grid cell present in the feature map. Aggregation of these results gives the final result.

We fine-tuned a pre-trained YOLOv8 model using our paddy disease dataset, which consists of annotated images with bounding boxes around affected areas. This process involves transferring the knowledge from the pre-trained model and adapting it to the specific characteristics of paddy disease symptoms. The model is trained using a combination of three kinds of loss functions which include Bounding Box Regression Loss, Confidence Loss, and Classification Loss, optimizing its performance in detecting and classifying various paddy diseases. Bounding Box Regression loss compares the bounding box coordinates Confidence Loss uses binary cross entropy loss, penalizing incorrect predictions about the presence or absence of objects inside a predicted bounding box. Lastly, classification loss evaluates the accuracy of the predicted class probabilities for each detected object, typically using a categorical cross-entropy loss for multi-class classification. The model's performance is evaluated using Mean Average Precision at 50% Intersection over Union (mAP50), ensuring it meets the high standards required for practical agricultural applications.

# Experiments

Our study involved conducting a series of experiments focusing on two primary tasks: classification and detection.

For classification purposes, CNN architectures including ResNet, VGG, MobileNet, and the Vision Transformer (ViT) were used as the deep learning models. The experiments were conducted on Kaggle's platform using an NVIDIA Tesla P100 GPU with 16GB VRAM, and the PyTorch framework. TensorBoard was employed for tracking experimentation metrics and progress. All images were resized to either 256x256 or 384x384 pixels depending on the experiment, with pixel values normalized using the mean and standard deviation of the ImageNet dataset for each of the three channels.

The models were trained with a batch size of 16, and 100 epochs. The learning rate started higher at 0.01 in the initial experiments and gradually decreased to as low as 1e-4 depending on the specific experiment. All the models were trained with Adam Optimizer. Each model, pre-trained on ImageNet, was fine-tuned on our dataset to leverage its unique strengths, such as hierarchical feature extraction in ResNet, detailed spatial features in VGG, and efficient depth-wise separable convolutions in MobileNet. and global self-attention mechanism in ViT. Model performance was evaluated using accuracy, precision, recall, F1-score, and confusion matrix metrics to ensure high standards of accuracy and reliability in real-world paddy disease identification scenarios.

**Paddy Disease Detection**

For our disease detection task, the fine-tuning of a pre-trained YOLOv8 model was used. The experiments were conducted in the same environment settings as in classification tasks. We utilized the Ultralytics library for YOLOv8 implementation and TensorBoard for tracking experimentation metrics and progress.

The models were trained with a batch size of 16 and the AdamW optimizer, using an image input size of **640x640 pixels**, which strikes a balance between computational efficiency and detection accuracy. Hyperparameter tuning was carried out to search for the best set of hyperparameters, after which the model was trained for 300 epochs with the obtained set of best hyperparameters. The learning rate was adjusted as needed throughout the training process to optimize model performance. The YOLOv8 model's performance was evaluated using box precision and recall, as well as mean Average Precision at 50% Intersection over Union (mAP50). These metrics ensured that the model met high standards of accuracy and reliability for real-world paddy disease detection scenarios.

## Evaluation Metrics

The evaluation matrix accuracy, precision, recall, F1-score, and confusion matrix metrics for the classification task and box precision, box recall, and mean Average Precision at 50% Intersection over Union (mAP50) for the object detection task were used. These metrics ensured comprehensive performance assessment across both tasks.

**Accuracy**

Accuracy measures the proportion of correct predictions out of the total predictions, providing an overall performance indicator of the model.

$$Accuracy = \frac{TP + TN}{TP + TN + FP + FN} \dots\dots\dots[2]$$

**Precision**

Precision indicates the proportion of true positive predictions out of the total predicted positives, reflecting the model's ability to avoid false positives.

$$Precision = \frac{TP}{TP + FP} \dots\dots\dots[3]$$

**Recall**

Recall, also known as sensitivity, measures the proportion of true positive predictions out of the actual positives, indicating the model's ability to capture all relevant instances.

$$Recall = \frac{TP}{TP + FN} \dots\dots\dots[4]$$

**F1-Score**

F1-score is the harmonic mean of precision and recall, providing a single metric that balances the trade-off between precision and recall.

$$F1 - Score = 2.\frac{Precision.Recall}{Precision + Recall} \dots\dots\dots[5]$$

**Confusion Matrix**

The confusion matrix is a table that summarizes the performance of a classification model by showing the counts of true positives (TP), true negatives (TN), false positives (FP), and false negatives (FN). It provides detailed insights into the types of errors the model is making, such as distinguishing between different classes. It is a vital tool in our study as many diseases are similar looking and can be misclassified as one another, highlighting areas where the model may need improvement.

**Box Precision and Box Recall**

These metrics evaluate how accurately the model localizes disease-affected areas in paddy leaves using bounding boxes and its ability to detect all instances of diseases despite variations in their appearance.

**Mean Average Precision at 50% Intersection over Union (mAP50)**

It is the average precision across all classes with a threshold of 50% overlap between the predicted and ground truth bounding boxes, providing a comprehensive measure of the model's detection performance. This metric was used to evaluate our model's performance in object detection tasks, ensuring that both precision and recall were properly assessed across all classes. The equation for calculating mean average precision (mAP) is defined in equation 6.

$$mAP50 = \frac{1}{N}\sum_{i=1}^{N} AP_i \quad \text{...............................................[6]}$$

Where, $AP_i$ is the average precision of class i and N is the number of classes.

## Mobile application development

A smartphone application was designed and developed to implement the paddy disease detection and classification task. It communicates with the web server connected to various backend microservices like DBMS, hosted ML servers, and messaging servers. Distributed databases and App scale sets through Azure services were used as the fault tolerant. Cloud services like Azure Virtual Networks, VM scale sets, and Azure DNS were used.

**System Design**

The overall design of the system is based on a distributed architecture comprising five major services.

**Backend web server**

Our backend server was based on Echo, a Golang web framework. The Golang is simple and has good performance, and single binary or no dependency properties so that it becomes highly portable and contain sizable. Entgo ORM(Object-Relational Mapper) is used to communicate with our database. This ORM is easy to set up, define relationships with, and query data across tables using edges and nodes i.e. graph-like properties. It handled some typical web server jobs and responsibilities like user authentication, image uploads, dispatching actions to RabbitMQ, retrieving inference results from RabbitMQ, and such business logic.

**Machine Learning server**

ML server was based on Python and its only job was to take an image input and produce inference results. It was based on master-worker architecture where one master monitors, spawns, and dismisses multiple workers. Workers are where real inference happens. The main objective of this type of architecture is to utilize the processing units to their maximum limit using threaded execution. Another objective was too hot-swap models or utilize multiple inference models using workers. There were two inference models abstracted as workers. Depending on whether we are performing classification or detection tasks, we can use corresponding workers to do that on runtime.

**Messaging Server for inter-service communication**

RabbitMQ was used to bridge our two types of servers i.e. ML server and Web server. The main objective of separating these servers and bridging with a queue service was to support horizontal scalability and separating concerns. In the case of scalability, we can independently, according to usage, increase or decrease these servers ensuring the reliability of our services. While separation of concern was important to remove dependencies. Using these queuing services also helps in load balancing which is crucial to distribute heavy inference tasks to an equal amount of available computing servers. RabbitMQ as a queuing service was used, among others like Kafka, because of its popularity, stability, and nothing else.

**PostgreSQL Database Server**

Being the most popular, stable, performant, and extensible RDBMS, we had to choose this as our primary database. We currently don't have many complicated relations, but if we had one, we would not be at a disadvantage due to the database.

**Mobile Application**

We developed a mobile application as the primary way to consume these services because of their availability and ease of use. Mobile phones have become the most penetrated electronic devices in the world with over 4.4 billion people using the mobile internet. Other reasons for using mobile apps as the primary consuming source are: portability, access to camera features, mobile internet accessibility, and ease of use.

ReactNative is used to develop this application because of its portable nature, rich tooling support, rapid development curve, and familiarity with JavaScript. Some of the features of this application are:

- User authentication to bind inference results to users
- Upload diseased plant pictures
- Display inference progress and results
- Display globally mapped disease outbreaks
- Suggest solutions to identified diseases.

All of these resources were collaboratively worked on GitHub and deployed in Azure with the help of Terraform, a cloud infrastructure management service, keeping the horizontal scalability of the system in mind.

During deployment altogether forteen individual resources were managed in Azure. Some examples are VMs, Virtual Networks, persistent logging disks, a basic security group to protect deployed machines from public access, azure managed Postgresql database server, global private DNS for them to contact each other easily and efficiently, and a public IP to access these services. Note that only one public IP to expose necessary ports like SSH, HTTP, etc. to the public internet was used. If we needed to maintain private services, we would need to access them through this one single machine. The complete dataflow diagram is shown in Figure 7.

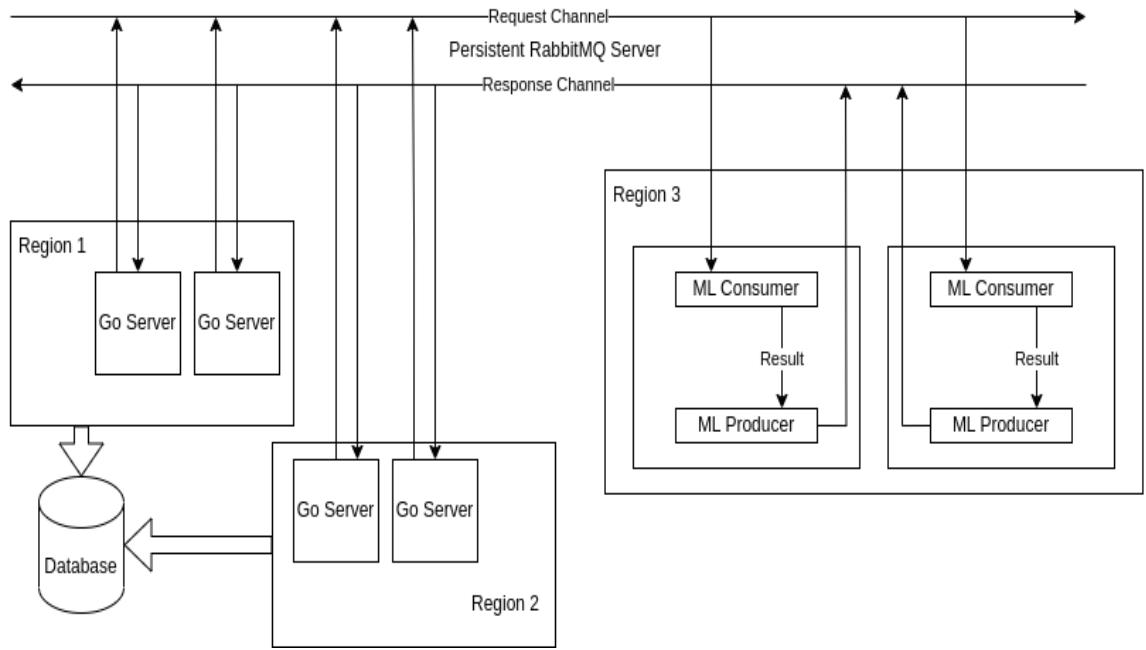

Figure 7: System Diagram of Mobile Application Backend

The one main way to interact with this system is with a mobile application. Figure 8 represents a simple flowchart to represent a basic flow of how a user goes through to use this system.

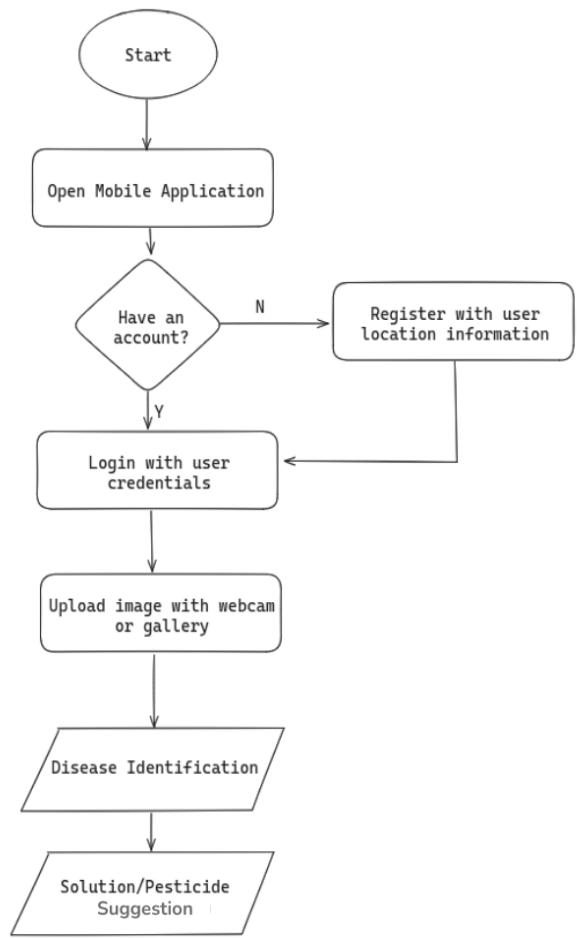

Figure 8: Workflow of the proposed system

The user needs to have a basic account set up first. Upon setting that, they can upload a photo through our photo section. After a couple of seconds of analyzing the image, our server will spit out a detection result that displays which sections of that plant have diseases, what type of disease they are, and suggest methods to help remedy them.

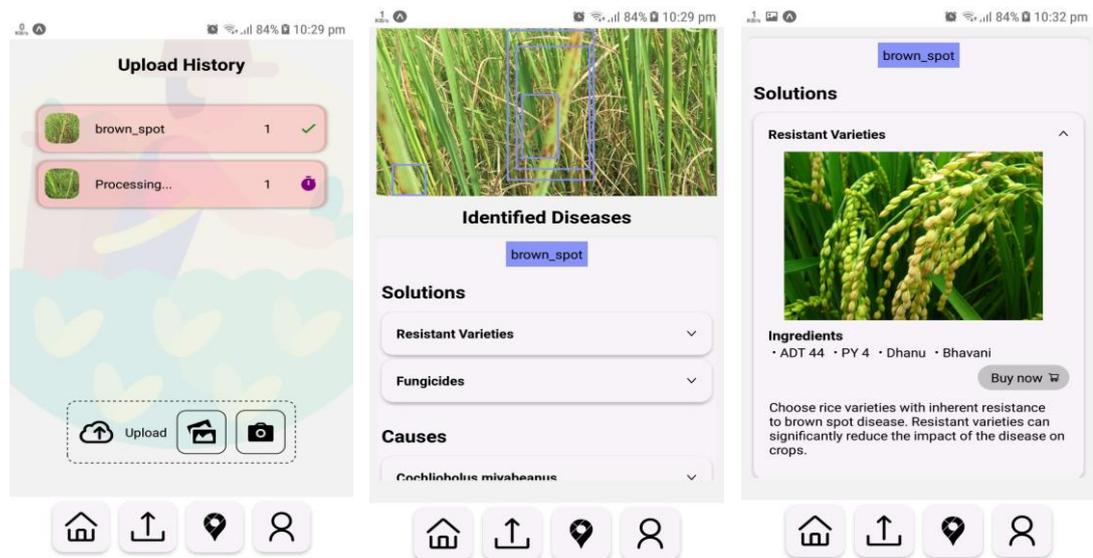

Figure 9: Mobile Application User Interface Overview

## Result and Discussion

Several CNN and ViT architectures were experimented with and best best-performing models in the initial experiments were manually hype tuned and trained further. Experimental results of some of the best-performing models are summarized in Table 1.

Table 1: Experimental results of paddy disease classification task

| Experiment name | Val Acc | Val F1 |
| --- | --- | --- |
| ResNet18 | 85.67 | 85.21 |
| VGGNet16 | 86.14 | 85.48 |
| ResNet50 | 92.4 | 92.4 |
| MobileNet | 94.8 | 94.2 |
| ResNet50 | 97.6 | 97.4 |
| ViT-16x16 | 98.3 | 98.3 |

Resnet50 was found to be the best performing CNN model whereas Vision Transformer with 16x16 patch performed best for the transformer model.

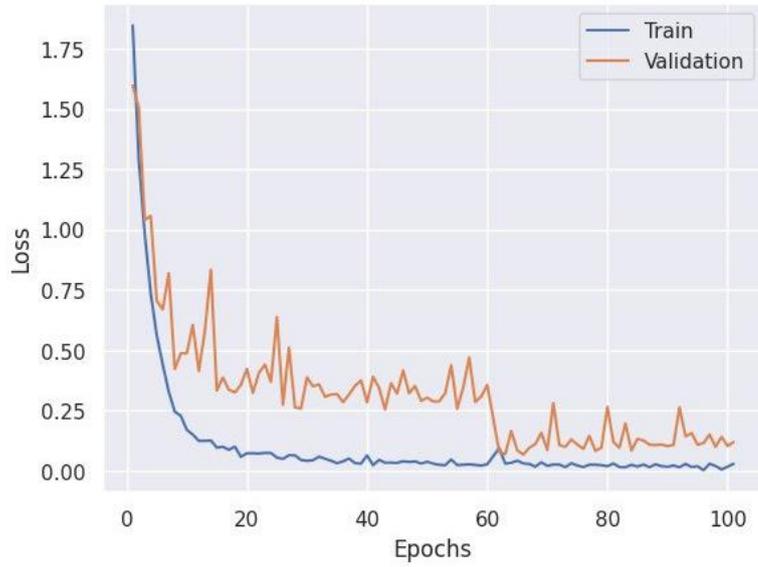

Figure 10: Loss Comparison for Vision Transformer Between Training and Validation

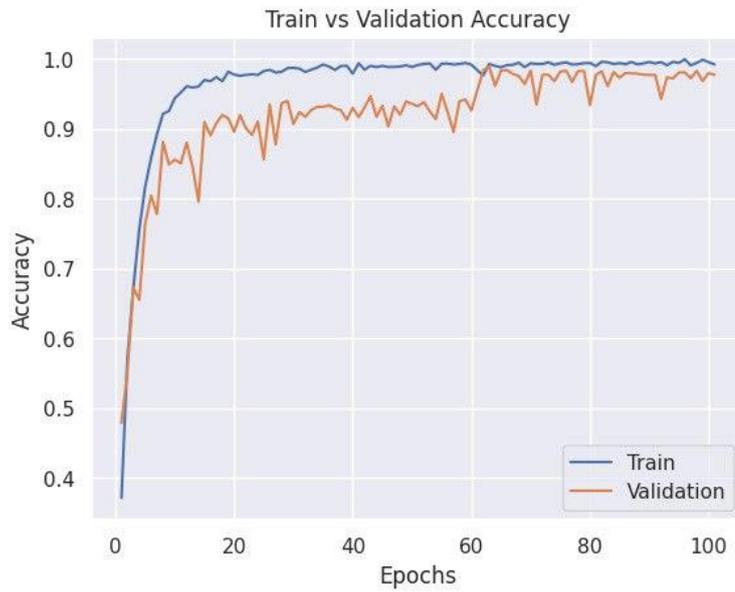

Figure 11: Accuracy Comparison of Vision Transformer Between Training and Validation

For the detection task, a hyperparameter search was conducted by training for 20 epochs on each set. The best hyperparameters were then used to train the model for 300 epochs.

Table 2: Experimental results of detection task

| Class | Box Precision | Box Recall | map50 |
| --- | --- | --- | --- |
| all | 73.6 | 65.7 | 69.0 |
| bacterial_leaf_blight | 72.8 | 52.8 | 50.9 |
| bacterial_leaf_streak | 87.4 | 90.4 | 89.1 |
| bacterial_panicle_blight | 71.2 | 76.5 | 78.6 |
| black_stem_borer | 73.4 | 78.8 | 75.3 |
| blast | 84.0 | 58.2 | 66.4 |
| brown_spot | 77.6 | 42.1 | 55.4 |
| *downy_mildew* | *68.4* | *59.6* | *67.3* |
| *hispa* | *35.4* | *21.5* | *21.9* |
| *leaf_roller* | *84.8* | *71.9* | *76.2* |
| *tungro* | *75.0* | *81.8* | *82.4* |
| *white_stem_borer* | *76.2* | *82.8* | *83.8* |
| *yellow_stem_borer* | *77.5* | *72.4* | *80.6* |

These experiment results of the Yolov8 detection model reveal varying performance across different disease classes. With most of the disease classes achieving high mean average precision, few classes like bacterial leaf blight, brown spot, and hispa indicate areas of improvement. The three lowest-performing diseases have smaller disease patches, which the model finds challenging to detect accurately.

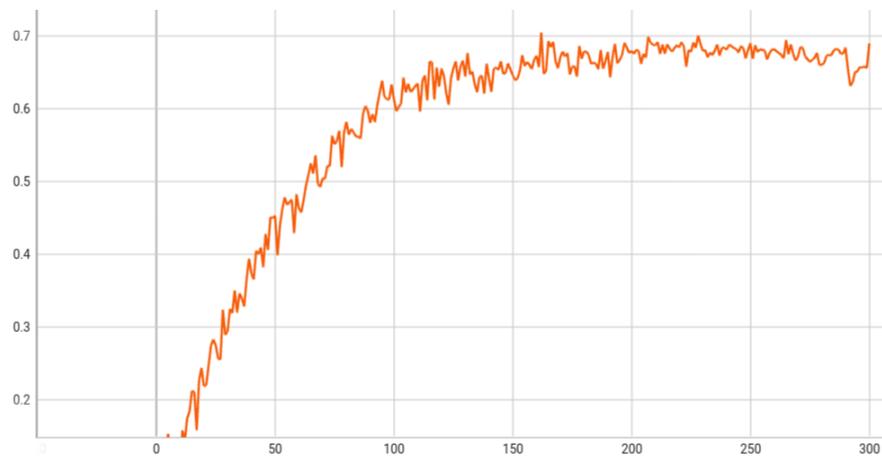

Figure 12: mAP50 Evolution During Training for Detection Task

In this study, two major tasks: classification detection were performed to explore the approaches to identify various paddy diseases using deep learning. For classification, different CNN architectures and vision transformers were explored. Resnet50 was found to be the best-performing CNN model with 97.6% validation accuracy while Vision Transformer with a 16x16 patch slightly outperformed this CNN model with a validation accuracy of 98.7%. Similarly, for detection tasks, the yolov8 model was experimented with various hyperparameters and achieved the mAP50 of 69.%. The inference time for Resnet50, Vision Transformer, and yolov8 models were 650 ms, 2.4s, and 472 ms respectively.

Though the classification models were more effective in accurately identifying paddy disease, they were unable to work with images with varying zoom scales and also could not identify multiple disease instances at once. Whereas the yolov8 model could identify paddy disease at different scales and could identify multiple diseases at once at the cost of some accuracy. To utilize the strength of both these approaches, a mobile application was developed where the yolov8 model was used for initial detection, and the classification model was used for further verification from a remote server.

# Conclusion

Overall, this study highlights the potential of deep learning models in paddy disease identification and demonstrates how combining classification and detection models may produce a reliable solution for practical agricultural applications. Leveraging the strength of advanced architectures like CNN, Vision Transformer, and YOLO, we have shown the possibility of using an AI-driven agricultural diagnosis system for different plant diseases. To ensure wider impact and adoption among farmers, future research and development should concentrate on improving these models using a more diverse set of training images covering more diseases and broadening their applicability. Enhancing mobile applications with features like a better treatment recommendation system, an interactive chatbot-based interface for real-time farmer support, and local agricultural advisory systems should also be carried out. Farmers in resource-constrained regions would find the applications more useful, accessible, and efficient because of these developments.